\documentclass[twocolumn,times,final]{elsarticle}
\usepackage{framed,multirow}

\usepackage{amssymb}
\usepackage{latexsym}

\usepackage{url}
\usepackage{xcolor}
\definecolor{newcolor}{rgb}{.8,.349,.1}

\usepackage{times}
\usepackage{soul}
\usepackage{url}
\usepackage[hidelinks]{hyperref}
\usepackage[utf8]{inputenc}
\usepackage[small]{caption}
\usepackage{graphicx}
\usepackage{amsmath}

\usepackage{booktabs}
\usepackage{algorithm}
\usepackage{algorithmic}
\usepackage[switch]{lineno}

\usepackage{verbatim}
\usepackage{subfigure}
\usepackage{amsfonts}
\usepackage{lipsum}
\usepackage{makecell}
\journal{arXiv}

\begin{document}

\thispagestyle{empty}

\clearpage

\setcounter{page}{1}

\begin{frontmatter}

\title{Causal Interpretability for Adversarial Robustness: A Hybrid Generative Classification Approach}

\author[1]{Chunheng Zhao} 
\author[1]{Pierluigi Pisu\corref{cor1}}
\cortext[cor1]{Corresponding author: 
  Tel.: +1-864-283-7227;  
  fax: +1-864-283-7208;}
\ead{pisup@clemson.edu}
\author[2]{Gurcan Comert}
\author[2]{Negash Begashaw}
\author[3]{Varghese Vaidyan}
\author[4]{Nina Christine Hubig}

\date{}


\affiliation[1]{organization={Department of Automotive Engineering, Clemson University},
                addressline={4 Research Drive}, 
                city={Greenville}, 
                postcode={29607}, 
                state={South Carolina},
                country={USA}}

\affiliation[2]{organization={Department of Computer Science, Physics \& Engineering, Benedict College},
                addressline={1600 Harden Street}, 
                city={Columbia}, 
                citysep={},
                postcode={29204}, 
                state={South Carolina},
                country={USA}}

\affiliation[3]{organization={Department of Computer and Cyber Sciences, Dakota State University},
                addressline={820 N Washington Avenue}, 
                city={Madison}, 
                citysep={},
                postcode={57042}, 
                state={South Dakota},
                country={USA}}

\affiliation[4]{organization={School of Computing, Clemson University},
                addressline={821 McMillan Road}, 
                city={Clemson}, 
                postcode={29634}, 
                state={South Carolina},
                country={USA}}


\begin{abstract}

Deep learning-based discriminative classifiers, despite their remarkable success, remain vulnerable to adversarial examples that can mislead model predictions. While adversarial training can enhance robustness, it fails to address the intrinsic vulnerability stemming from the opaque nature of these black-box models. We present a deep ensemble model that combines discriminative features with generative models to achieve both high accuracy and adversarial robustness. Our approach integrates a bottom-level pre-trained discriminative network for feature extraction with a top-level generative classification network that models adversarial input distributions through a deep latent variable model. Using variational Bayes, our model achieves superior robustness against white-box adversarial attacks without adversarial training. Extensive experiments on CIFAR-10 and CIFAR-100 demonstrate our model's superior adversarial robustness. Through evaluations using counterfactual metrics and feature interaction-based metrics, we establish correlations between model interpretability and adversarial robustness. Additionally, preliminary results on Tiny-ImageNet validate our approach's scalability to more complex datasets, offering a practical solution for developing robust image classification models.

\end{abstract}


\end{frontmatter}


\section{Introduction}
In recent years, various deep learning-based discriminative models have made significant progress in image classification \cite{mauricio2023comparing, wang2021comparative}, such as VGG \cite{simonyan2014very}, ResNet \cite{he2016deep}, and ViT \cite{dosovitskiy2020image}.  Despite their great success, recent studies have uncovered that deep discriminative classifiers are vulnerable to adversarial examples \cite{szegedy2013intriguing,goodfellow2014explaining,carlini2017towards,madry2017towards}. These are well-designed inputs with slight perturbations that can cause the model to produce incorrect classification results. This vulnerability partly stems from the opaque nature of the black-box deep neural network (DNN) models and their limited interpretability. Adversarial examples are typically modified slightly so that humans don't misclassify them, thus posing a non-trivial threat to many DNN-based applications.

Adversarial training has emerged as one of the most effective techniques for improving robustness \cite{goodfellow2014explaining,shafahi2019adversarial,wong2020fast}. The core idea is to include adversarial examples in the training stage. However, training data augmentation doesn't address the root cause of adversarial vulnerability: the lack of model interpretability. Recent studies show that deep generative classifiers, despite their limited performance on image classification tasks \cite{mackowiak2021generative, fetaya2019understanding, zheng2023revisiting}, can be more robust to adversarial examples \cite{li2019generative}. This adversarial robustness benefits from their ability to model the distribution of each class, essentially modeling how a specific class generates the input data. This approach leads to better model interpretability and allows them to better understand what adversarial inputs might look like. Discriminative classifiers, on the other hand, model decision boundaries between different classes and learn which input features contribute most to distinguishing between various classes. Consequently, they are more vulnerable to adversarial examples designed to create outliers for decision boundaries, thus confusing the classifier. This difference provides the underlying motivation and theoretical support for using generative classifiers to build more adversarially robust networks, as it's more challenging to shift feature distribution than to create outliers.

Building upon these insights, in this paper, we develop a generalized deep neural network architecture for image classification: an ensemble network consisting of discriminative features and generative classifiers. This novel architecture combines the strengths of both approaches to create an accurate and more robust classification model. Our methodology begins with constructing a latent variable model that models the relationships among input images, discriminative features, output labels, and latent variables. We then employ variational Bayes to formulate final prediction probabilities. Our contributions in this work can be summarized as follows: (1) We propose a bottom-up discriminative-generative architecture featuring a generative component that can be attached to various pre-trained discriminative models, highlighting the generalizability of our approach; (2) We demonstrate our network's resistance to various types and strengths of white-box adversarial attacks, significantly decreasing attack success rates without adversarial training; (3) Leveraging counterfactual metrics and feature interaction-based metrics, we demonstrate that the proposed model has superior model interpretability.

\section{Related Work}
\label{2}

Adversarial training has emerged as an effective approach to enhance model robustness. Goodfellow et al. \cite{goodfellow2014explaining} and Kurakin et al. \cite{kurakin2016adversarial} demonstrated significant error rate reductions on MNIST and ImageNet datasets, respectively, through adversarial training. However, injecting adversarial examples into the training set can decrease accuracy on clean datasets, and the process is computationally expensive due to the construction of sophisticated adversarial examples \cite{kurakin2016adversarial}. To address these issues, Shafahi et al. \cite{shafahi2019adversarial} introduced a "free" adversarial training algorithm that eliminates the overhead cost of generating adversarial examples by recycling the gradient information computed when updating model parameters. Recent improvements in adversarial training include using cyclic learning rates to accelerate training \cite{wong2020fast} and employing diffusion models to augment adversarial training datasets \cite{wang2023better}. Nevertheless, it remains unrealistic to involve all types of adversarial examples in the training phase due to the variety of attacks. More importantly, adversarial training doesn't solve the intrinsic problem of adversarial vulnerability, which lies in the lack of model interpretability.

Recent research has begun to evaluate the adversarial robustness of generative classifiers, which offer improved interpretability. Li et al. \cite{li2019generative} found that deep generative classifiers can be robust on MNIST and CIFAR-binary datasets. Zhang et al. \cite{zhang2020causal} further improved deep generative classifier robustness by modeling adversarial perturbation from a causal perspective. However, both studies were unable to obtain satisfactory results on full image datasets, reporting that VAE-based generative classifiers achieved less than 50\% clean test accuracy on CIFAR-10. Additionally, they didn't conduct an interpretability analysis to show the correlations between adversarial robustness and model interpretability. Mackowiak et al. \cite{mackowiak2021generative} showed that Invertible Neural Networks (INNs) are more interpretable than ResNet-50. However, they demonstrated that INNs do not fully prevent C\&W attacks \cite{carlini2017towards} on ImageNet, despite C\&W attacks prioritizing small perturbations over attack success rates.

To address these limitations, we develop a generalized deep generative classifier architecture for more complex datasets (full CIFAR-10 and CIFAR-100), using only clean data for training to reduce training time and computational costs. To demonstrate the generalizability of our approach, we conduct preliminary experiments on the Tiny-ImageNet dataset, showcasing the potential of our method to scale to larger datasets for real-world applications. Furthermore, we evaluate robustness against one of the strongest attacks (PGD \cite{madry2017towards}), which prioritizes attack success rates over perturbation size. Importantly, we explore the correlations between adversarial robustness and model interpretability, showing that by designing causal graphs in generative classifiers, we can achieve better model interpretability, leading to improved adversarial robustness.

\begin{figure}
  \centering
  \includegraphics[width=0.9\linewidth]{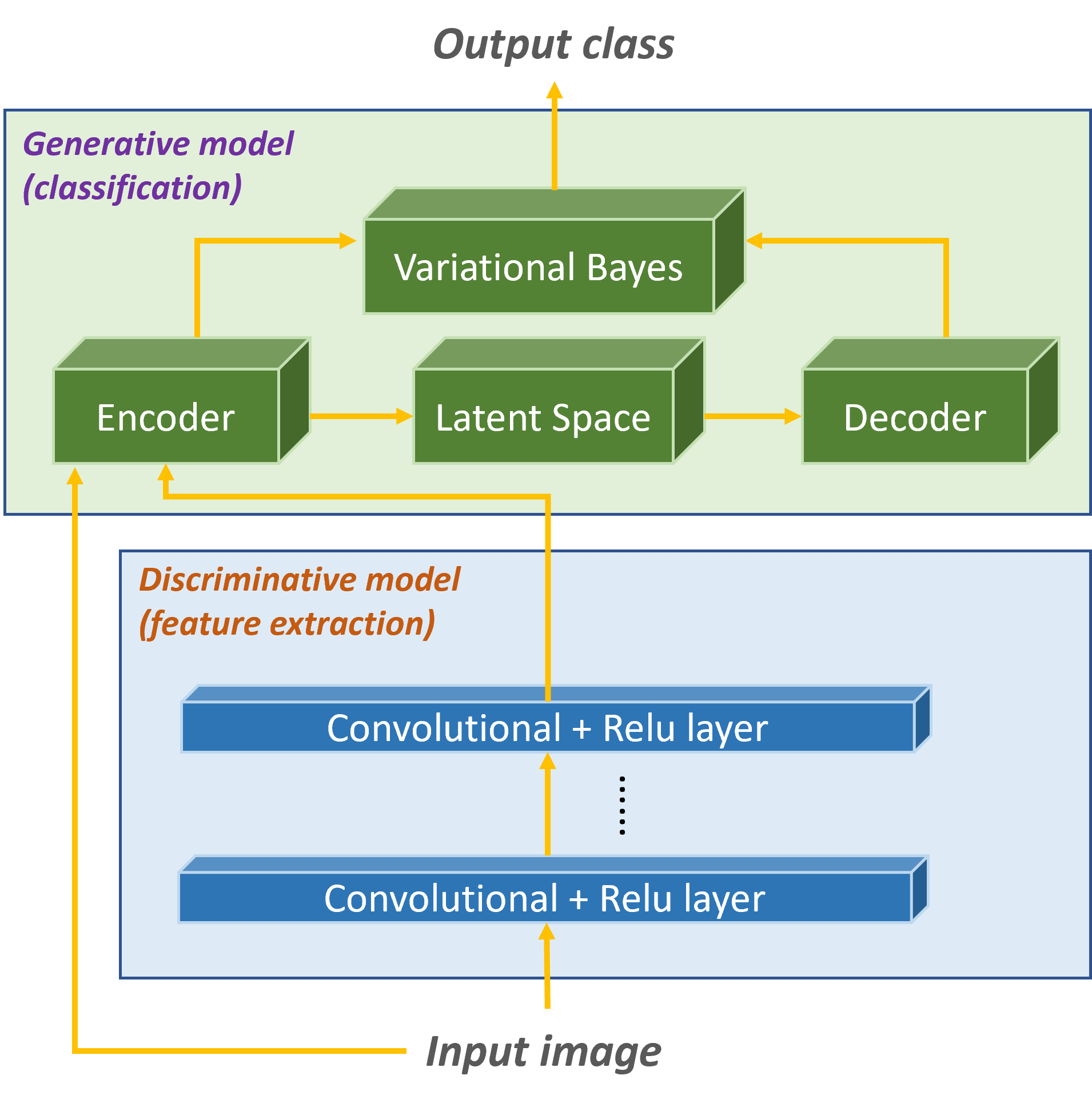}
  \caption{Bottom-up discriminative generative architecture. The overall model consists of both a feature extractor and a generative classifier.}
  \label{arch}
\end{figure}

\section{Methodology}
\label{3}
The overall model architecture, illustrated in Fig.~\ref{arch}, consists of two primary components: a bottom-level pre-trained discriminative feature extraction network and a top-level generative classification network. While the discriminative features ensure high classification accuracy, the generative model provides adversarial robustness by modeling the distribution of adversarial inputs. The bottom-level discriminative network can be any pre-trained Convolutional Neural Network (CNN) model, such as VGG or ResNet, making our approach generalizable to various classification tasks and user preferences.  

Our work extends the approach presented in \cite{zhang2020causal} to handle more complex image classification tasks (e.g. full CIFAR-10, CIFAR-100 and Tiny-ImageNet datasets). The top-level generative classifier processes both the features extracted from the bottom-level pre-trained CNN and the original image. This dual-input approach leverages the generative model's ability to regenerate its inputs, enabling it to learn distributions from both feature representations and original data. Such comprehensive learning strategy enhances the model's capability to handle both adversarial images and adversarial features.

The generative component of our architecture consists of two main elements:
(1) A latent variable model captures the relationships between input images, discriminative features, output labels, and latent variables;
(2) A variational auto-encoder (VAE) formulates final prediction probabilities using the latent variable model and variational Bayesian inference.

\subsection{Latent Variable Model}
\label{modeling}

\begin{figure}[h!]
    \centering
    \includegraphics[width=0.9\linewidth]{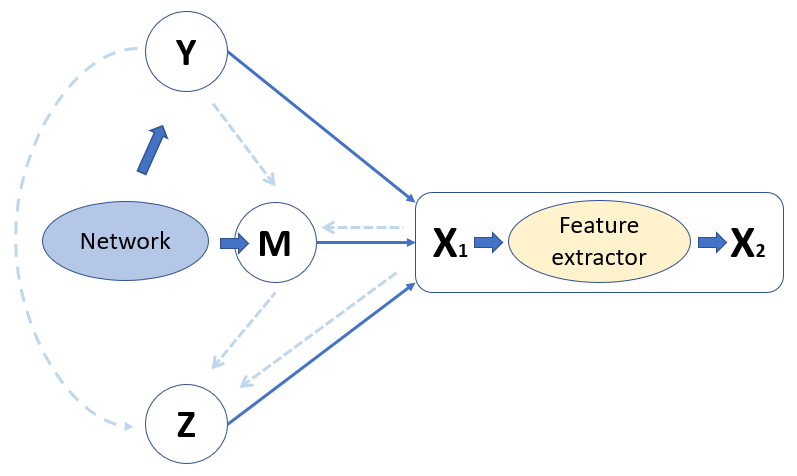}
    \caption{Causal graph. Solid lines represent the causal reasoning of input data.}
    \label{causal graph}
\end{figure}

The foundation of our generative classifier is a deep latent variable model with a causal graph that captures the relationships between inputs, outputs, and latent variables, as depicted in Fig.~\ref{causal graph}. This causal reasoning approach enables DNNs to learn causal relations rather than merely statistical correlations between inputs and outputs, thereby improving robustness and reducing overfitting. In our model, we define two types of inputs to the generative classifier which are the original image $X_1$ and image features extracted from a pre-trained feature extractor $X_2$. The causal graph incorporates three key factors that influence the formation of image data: $Y$ represents the predicted label containing the class of the object; $M$ represents variables that can be modified artificially (adversarial perturbations); $Z$ denotes all the other contributing factors.

Following Goodfellow et al. \cite{goodfellow2014explaining}, we model adversarial perturbations $M$ as a specific type of noise affecting both $X_1$ and $X_2$. In white-box attacks, these perturbations $M$ are generated based on the labels $Y$, input data $(X_1,X_2)$, and network parameters $\theta$. The formation of adversarial inputs $(X_1,X_2)$ is then caused by the combination of adversarial perturbations $M$, labels $Y$, and other factors $Z$. For simplicity, we do not consider cases where other factors $Z$ influence labels $Y$. The causal model for input data formation can be expressed as:
\begin{equation}
    X_1, X_2 = P(M,Y,Z)
    \label{eq:xadv}
\end{equation}
where $P$ represents the process of input data formation. The generative classifier learns this causal relationship during the training phase and makes correct classifications based on its reasoning from these factors during the inference phase.

\subsection{VAE-based Generative Classifier}
\label{vae}

Building upon the causal relations established in Section~\ref{modeling}, we develop our generative classifier using amortized variational inference \cite{zhang2018advances}. The fundamental approach applies Bayes' theorem to estimate the conditional probability $p(y|x)$, where $y$ is the target label and $x$ is the input. A generative classifier predicts the label $y$ of an input $x$ according to:

\begin{equation}
p(y|x)=\frac{p(x|y)p(y)}{p(x)}=softmax_{c=1}^{C}[\log p(x,y_c)]
\label{eq1}
\end{equation}
where $C$ is the total number of classes and the likelihood function $\log p(x,y_c)$ is maximized during training. The prediction process involves computing the log-likelihood for each class $y=c$ followed by a softmax operation. After incorporating the latent variables $m$ (adversarial perturbations) and $z$ (other factors), Eq.~(\ref{eq1}) can be reformulated as:

\begin{equation}
p(y|x)=softmax_{c=1}^{C}[\log \int p(x,y_c,z,m)\,dm \,dz]
\label{eq2}
\end{equation}

Given that our input $X$ contains both $x_1$ and $x_2$, we can express the joint probability $p(x,y_c,z,m)$ as:

\begin{equation}
\begin{split}
p(x_1, x_2,y,z,m) &= p(x_1,x_2|y,z,m)p(y,z,m) \\
&=p(x_1,x_2|y,z,m)p(m)p(z)p(y)
\end{split}
\label{eq3}
\end{equation}

Substituting Eq.~(\ref{eq3}) into Eq.~(\ref{eq2}) yields:

\begin{equation}
\begin{split}
&p(y|x)=softmax_{c=1}^{C} \\
&[\log \int p(x_1,x_2|y,z,m)p(m)p(z)p(y)\,dm \,dz]
\end{split}
\label{eq4}
\end{equation}

Due to the intractability of the integral of the marginal log-likelihood, arising from intractable true posterior densities $p(z|\cdot)$ and $p(m|\cdot)$ for latent variables, we introduce an approximate distribution $q(z,m;\lambda)$ with variational parameters $\lambda$ to approximate the true posterior \cite{kingma2013auto}. Then the training objective of maximizing log-likelihood function in Eq.~(\ref{eq4}) is equivalent to minimizing the Kullback-Leibler (KL) divergence \cite{kullback1951information} between the variational distribution and true posterior $D_{KL}(q(z,m;\lambda)||p(z,m|\cdot))$. However, this divergence $D_{KL}$ is almost impossible to minimize to zero because the variational distribution is usually not capable enough to catch the complexity of the true posterior due to insufficient parameters. To address this challenge, we maximize the Evidence Lower Bound (ELBO), which is equivalent to minimizing the divergence \cite{zhang2018advances}. ELBO is a lower bound on the log marginal probability of the data and can be derived from Eq.~(\ref{eq4}) using Jensen’s inequality:

\begin{equation}
\begin{split}
&\log p(x,y) = \log \int p(x_1,x_2|y,z,m)p(m)p(z)p(y)\,dm \,dz \\
&=\log \int p(x_1,x_2|y,z,m)p(m)p(z)p(y)\frac{q(z,m;\lambda)}{q(z,m;\lambda)} \,dm \,dz \\
&=\log \mathbb{E}_{q(z,m;\lambda)}\left[\frac{p(x_1,x_2|y,z,m)p(m)p(z)p(y)}{q(z,m;\lambda)}\right] \\
&\ge \mathbb{E}_{q(z,m;\lambda)}\left[\log \frac{p(x_1,x_2|y,z,m)p(m)p(z)p(y)}{q(z,m;\lambda)}\right]
\end{split}
\label{eq5}
\end{equation}

\begin{figure}[h!]
    \centering
    \includegraphics[width=1\linewidth]{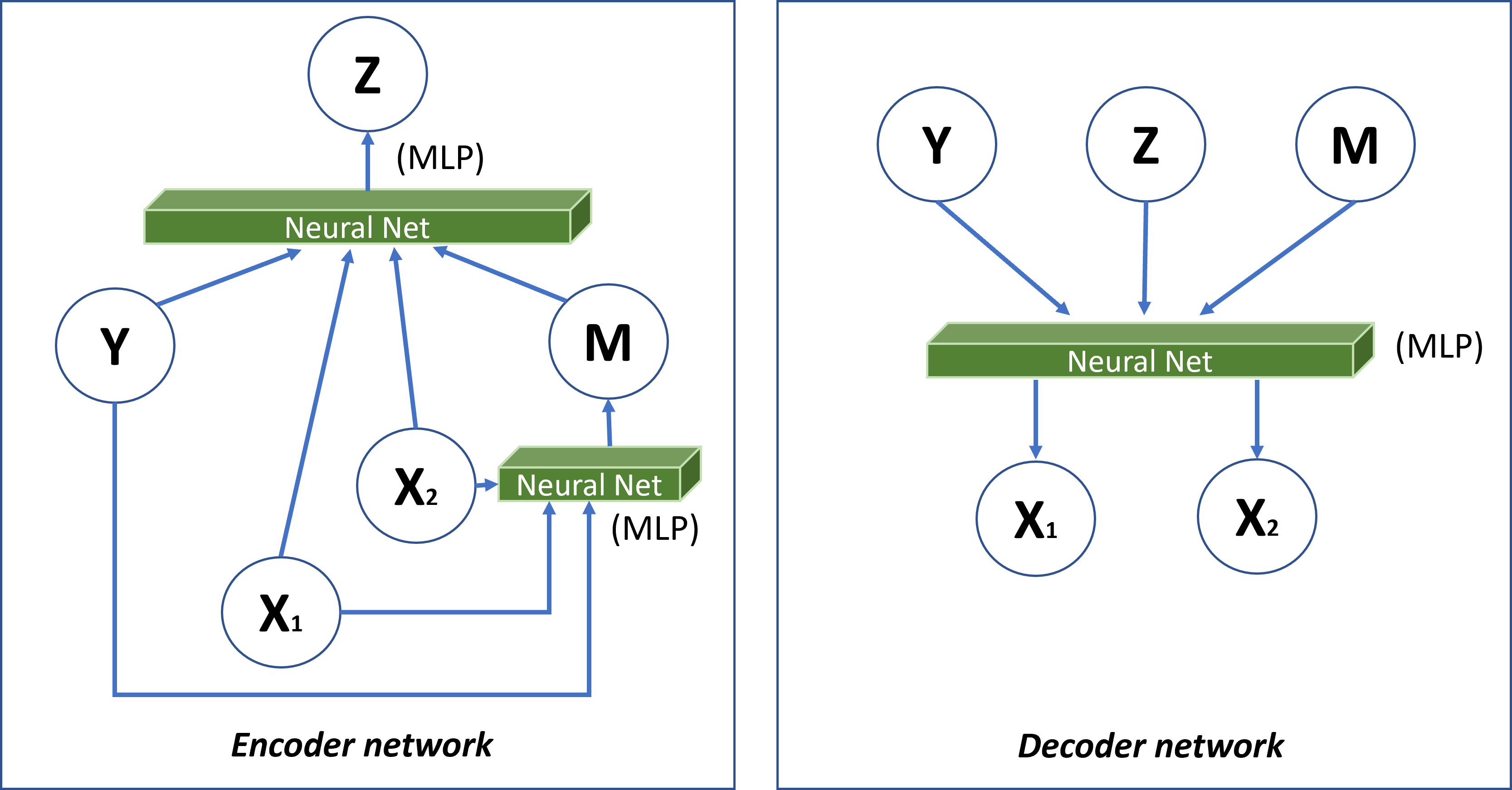}
    \caption{VAE architecture. Each individual neural net in the encoder and decoder estimates independent probabilities for $q$ and $p$, respectively.}
     \label{architecture}
\end{figure}

Following the causal graph (Fig.~\ref{causal graph}), we design the probabilistic encoder network (i.e., variational posterior inference) as:
\begin{equation}
\begin{split}
q(z,m;\lambda) &= q_{\lambda}(z,m|x_1,x_2,y) \\
&= q_{\lambda_1}(z|x_1,x_2,y,m)q_{\lambda_2}(m|x_1,x_2,y)
\end{split}
\label{eq6}
\end{equation}
where $\lambda = \{\lambda_1,\lambda_2\}$ are the encoder network parameters. $\lambda_1$ is parameter for encoder network $q_{\lambda_1}(z|x_1,x_2,y,m)$, and $\lambda_2$ is parameter for encoder network $q_{\lambda_2}(m|x_1,x_2,y)$. The generative parameters of the probabilistic decoder network are defined as:
\begin{equation}
p_{\theta}(x_1, x_2,y,z,m)=p_{\theta_1}(x_1,x_2|y,z,m)p(m)p(z)p(y)
\label{eq7}
\end{equation}
where the generative parameters are $\theta= \{\theta_1\}$ and $\theta_1$ is parameter for decoder network $p_{\theta_1}(x_1,x_2|y,z,m)$. The VAE architecture (Fig.~\ref{architecture}) implements both encoder and decoder networks with separate neural nets. Both probabilistic networks use multi-layer perceptrons (MLPs) with Gaussian outputs parameterized by mean $\mu$ and standard deviation $\sigma$. 


To avoid the enlarged dataset and computational overhead of adversarial training, we train only on clean data ($m=0$). Therefore, Eq.~(\ref{eq6}) and Eq.~(\ref{eq7}) can be simplified and the joint training of $p_{\theta}$ and $q_{\lambda}$ networks maximizes the simplified ELBO:
\begin{equation}
\max \mathbb{E}_{q_{\lambda_1}}\left[\log \frac{p(z)p(y_c)p_{\theta_1}(x_1,x_2|y,z,m)}{q_{\lambda_1}(z|x_1,x_2,y,m)}\right]
\label{train}
\end{equation}

We initialize $p(z)$ as a Gaussian distribution ($\mu = 0$, $\sigma = 0$) and $p(y)$ as a uniform distribution based on the dataset's classes (i.e., 0.1 for CIFAR-10 and 0.01 for CIFAR-100). During inference, $m$ is not set to $0$ but sampled from $q_{\lambda_2}(m|x_1,x_2,y_c)$ and $z$ is sampled from $q_{\lambda_1}(z|x_1,x_2,y_c,m)$. The final prediction is:
\begin{equation}
\begin{split}
&p(y|x)=\frac{p(x|y)p(y)}{p(x)} \\ 
&\simeq softmax^C_{c=1} \left[\log \sum_{k=1}^K \frac{p(z^k)p(y_c)p_{\theta_1}(x_1,x_2|y_c,z,m)}{q_{\lambda_1}(z^k|x_1,x_2,y_c,m)}\right]
\end{split}
\label{pred}
\end{equation}
where $K$ denotes the number of samples.

\section{Experiments}
\label{4}

\subsection{Experimental Setup}
\label{setup}
\noindent\textbf{Datasets.} Generative classifier benchmarking typically focuses on MNIST, SVHN, or CIFAR datasets due to performance constraints \cite{mackowiak2021generative, zheng2023revisiting}. In this paper, we utilize the full CIFAR-10 and CIFAR-100 \cite{krizhevsky2009learning} for comprehensive evaluation of both clean accuracy and adversarial robustness. Additionally, we conduct clean accuracy experiments on Tiny-ImageNet \cite{le2015tiny} to demonstrate the generalizability of our approach.

\noindent\textbf{Feature extractors.} We employ VGG-16 and VGG-19 \cite{simonyan2014very} as our pre-trained feature extractors. Despite not being the most recent models, VGG remains widely used as backbones in state-of-the-art (SOTA) CNN and transformer-based vision models \cite{hu2023vgg,amjoud2023object} and is capable enough to deal with CIFAR and Tiny-ImageNet data. Rather than using the entire VGG model, we extract features from intermediate layers. We implement a shallow model (containing only convolution layers) and a deep model (containing both convolution and dense layers) for ablation study, with further details discussed in Section~\ref{ablation}. Our bottom-up architecture design also allows seamless integration of alternative backbones, such as ResNet \cite{he2016deep}, without requiring modifications to the generative component.

\noindent\textbf{VAE-based classifier.} Each MLP estimating $q_{\lambda_1}(z|x_1,x_2,y,m)$ and $q_{\lambda_2}(m|x_1,x_2,y)$ consists of four hidden fully-connected layers with 500 neurons each and ReLU activation functions. MLP estimating $p_{\theta_1}(x_1,x_2|y,z,m)$ consists of two hidden layers. We train the model using Adam optimizer with a learning rate of $5\mathrm{e}{-5}$ and a batch size of 150 for 350 total training iterations.

\noindent\textbf{Adversarial attacks.} Following the evaluation framework in \cite{li2019generative, zhang2020causal}, we target white-box attacks and evaluate our model against Fast Gradient Sign Method (FGSM) and Projected Gradient Descent (PGD). FGSM \cite{goodfellow2014explaining} is selected for its simplicity and representativeness as a classic attack. PGD \cite{madry2017towards} represents SOTA attack performance and is consistently ranked among the strongest attacks \cite{xue2024diffusion,wang2020adversarial}. For both attacks, we evaluate perturbation magnitudes $\epsilon$ ranging from $0$ to $0.2$, where $\epsilon=0$ represents clean inputs. The perturbation size is constrained to maintain visual similarity between original and adversarial inputs. We employ the $l_2$-norm with $30$ steps for PGD attacks.

\noindent\textbf{Interpretability.} We use counterfactual explanations to evaluate interpretability, treating adversarial examples as a specific type of counterfactuals. We evaluate these based on proximity (perturbation size) and speed (attack iterations)  \cite{moraffah2020causal}. Lower proximity and higher speed indicate higher quality counterfactuals. Models more susceptible to generating high-quality counterfactuals demonstrate lower interpretability, and this indicates greater vulnerability to adversarial attacks. 

Additionally, we employ feature attribution analysis through Remove and Retrain (ROAR) \cite{hooker2019benchmark}. Using integrated gradients \cite{sundararajan2017axiomatic} for feature attribution, we replace 30\% of the most important pixels with the mean of CIFAR data, followed by model retraining and re-evaluation on the modified dataset. Greater accuracy degradation in this case indicates better interpretability, as it shows the model's reliance on more important features.

\begin{figure*}[h!]
    \centering
    \subfigure[Accuracy under FGSM]{
    \includegraphics[width=0.31\linewidth]{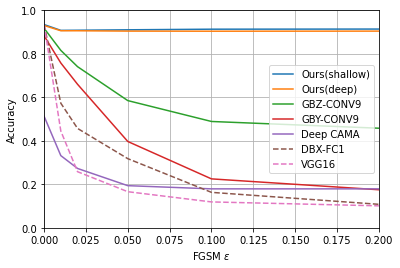}
    \label{FGSM}}
    \subfigure[Proximity under FGSM]{
    \includegraphics[width=0.31\linewidth]{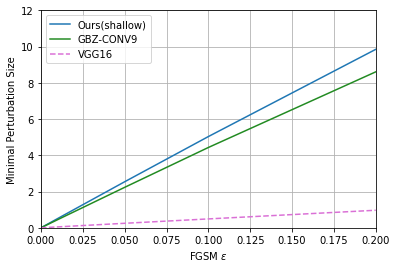}
    \label{FGSM1}}
    \subfigure[Speed under I-FGSM]{
    \includegraphics[width=0.31\linewidth]{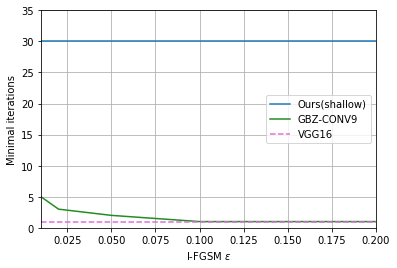}
    \label{FGSM2}}
    \subfigure[Accuracy under PGD]{
    \includegraphics[width=0.31\linewidth]{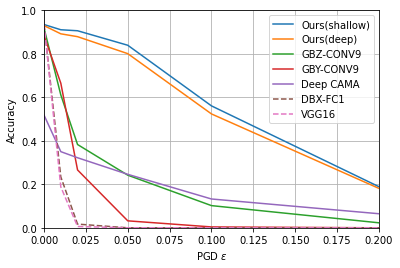}
    \label{PGD}}
    \subfigure[Proximity under PGD]{
    \includegraphics[width=0.31\linewidth]{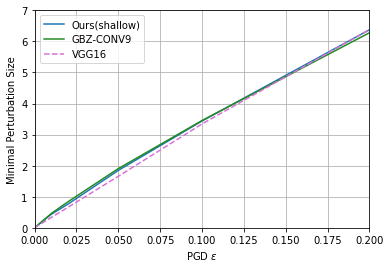}
    \label{PGD1}}
    \subfigure[Speed under PGD]{
    \includegraphics[width=0.31\linewidth]{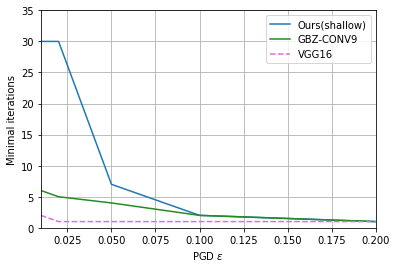}
    \label{PGD2}}
    \caption{Classification accuracy, adversarial example proximity, and adversarial example speed for FGSM and PGD on CIFAR-10 dataset. Dashed lines represent discriminative classifiers while solid lines represent generative classifiers. $\epsilon$ controls the attack strength.}
    \label{acc}
\end{figure*}

\begin{figure*}[h!]
    \centering
    \subfigure[Accuracy under FGSM]{
    \includegraphics[width=0.31\linewidth]{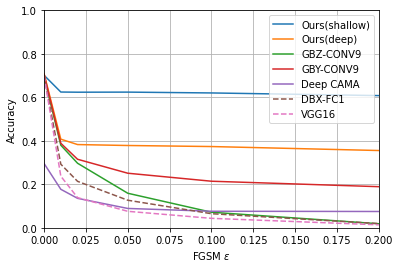}
    \label{FGSM-100}}
    \subfigure[Proximity under FGSM]{
    \includegraphics[width=0.31\linewidth]{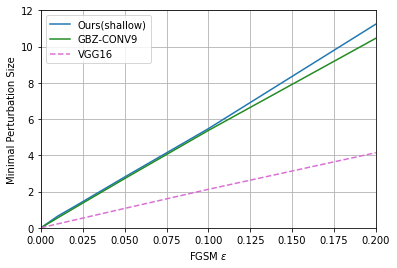}
    \label{FGSM1-100}}
    \subfigure[Speed under I-FGSM]{
    \includegraphics[width=0.31\linewidth]{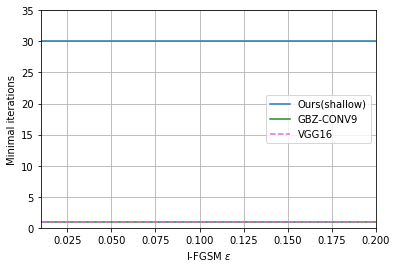}
    \label{FGSM2-100}}
    \subfigure[Accuracy under PGD]{
    \includegraphics[width=0.31\linewidth]{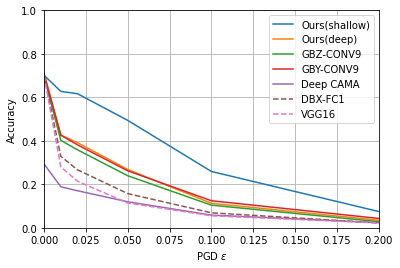}
    \label{PGD-100}}
    \subfigure[Proximity under PGD]{
    \includegraphics[width=0.31\linewidth]{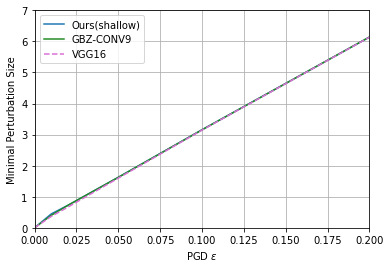}
    \label{PGD1-100}}
    \subfigure[Speed under PGD]{
    \includegraphics[width=0.31\linewidth]{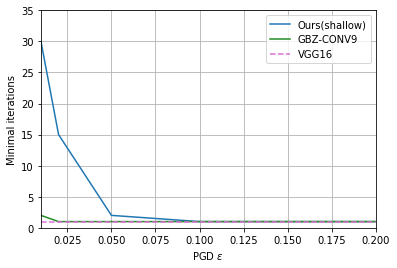}
    \label{PGD2-100}}
    \caption{Classification accuracy, adversarial example proximity, and adversarial example speed for FGSM and PGD on CIFAR-100 dataset. Dashed lines represent discriminative classifiers while solid lines represent generative classifiers. $\epsilon$ controls the attack strength.}
\end{figure*}

\subsection{Evaluation of Adversarial Robustness}
\label{robustness}

We evaluate our model's robustness against adversarial attacks as shown in Fig.~\ref{FGSM}, \ref{PGD}, \ref{FGSM-100}, and \ref{PGD-100}. The baseline models include three adversarially robustified generative classifiers (GBZ-CONV9 \cite{li2019generative}, GBY-CONV9 \cite{li2019generative}, and Deep CAMA \cite{zhang2020causal}) and one discriminative classifier (DBX-FC1 \cite{li2019generative}). Additionally, as our model employs VGG16 as a feature extractor, we include a pure VGG-based discriminative classifier for direct comparison. Results from both shallow and deep models are reported and ablation study is shown in Section~\ref{ablation}.

Our proposed model demonstrates superior adversarial robustness across different attack types and strengths. For FGSM attacks, the model maintains remarkable robustness on both CIFAR-10 and CIFAR-100 datasets, with accuracy degradation limited to less than $0.1$, effectively mitigating the FGSM attack's impact. Against the more challenging PGD attack, our model achieves impressive results on CIFAR-10, maintaining over $0.8$ accuracy at $\epsilon=0.05$ and $0.5$ accuracy at $\epsilon=0.1$, representing a significant robustness improvement. While the PGD attack results on CIFAR-100 show reduced effectiveness compared to CIFAR-10, our model still outperforms all baseline models. The generative classifiers GBZ-CONV9 and GBY-CONV9 also show robustness on both attacks and datasets, but the improvement is limited compared to our approach. Notably, Deep CAMA fails to generalize effectively to CIFAR-10 and CIFAR-100, achieving clean accuracy below $0.5$ on both datasets, consistent with the limitations reported in their paper \cite{zhang2020causal}. The discriminative baselines (VGG16 and DBX-FC1) exhibit the worst robustness across both attacks and datasets. Their performance degrades significantly even with minimal perturbations ($\epsilon < 0.05$), with accuracy falling below $0.2$. These results highlight the advantages of our hybrid approach over purely discriminative or generative classifiers.

Additionally, our model maintains high clean dataset accuracy ($\epsilon=0$) without the typical degradation observed in adversarial training \cite{kurakin2016adversarial}. We attribute this to the enhanced interpretability (showed in Section~\ref{caus}), which appears to contribute to overall model performance. Traditional adversarial training typically requires 3-30 times longer training periods compared to non-robust equivalents \cite{shafahi2019adversarial}. While our study doesn't include quantitative training time comparisons with adversarially trained baselines, our approach provides inherent efficiency advantages as it achieves robustness without dataset augmentation or enlargement through injecting adversarial examples.

\subsection{Evaluation of Model Interpretability}
\label{caus}

For a focused interpretability analysis, we compare our best-performing model (shallow model) against the most robust generative baseline, GBZ-CONV9, and a pure VGG16 discriminative model. Our evaluation employs multiple complementary metrics to assess model interpretability. We first examine counterfactual proximity by measuring the perturbation size needed in adversarial examples to affect model predictions. Models with better interpretability typically require larger perturbation (low proximity) for successful attacks. The FGSM attack results, presented in Fig.~\ref{FGSM1} and \ref{FGSM1-100}, demonstrate that both our model and GBZ-CONV9 require larger perturbation compared to VGG16 to achieve effective attacks. This increased perturbation requirement leads to low counterfactual proximity, indicating our model's enhanced interpretability. For PGD attacks (Fig.~\ref{PGD1} and \ref{PGD1-100}), the differences between models are less apparent, likely because the iterative nature of PGD allows perturbations to accumulate over multiple steps ($T=30$), potentially masking model-specific variations in perturbation size.

Our analysis of counterfactual generating speed reveals additional insights into model interpretability. We define this metric as the minimum number of iterations required to reduce accuracy to $0.4$, with slower attack speed (more iterations) indicating increased attack difficulty and better model interpretability. The results for Iterative-FGSM (I-FGSM) and PGD attacks, shown in Fig.~\ref{FGSM2} and \ref{FGSM2-100}, reveal significant differences between models. VGG16 proves highly vulnerable to single-step attacks, with FGSM and PGD attacks executing fast. GBZ-CONV9 shows moderate resistance, requiring $5$ iterations on CIFAR-10, though remaining vulnerable to single-step attacks on CIFAR-100. Our model demonstrates remarkable resistance to I-FGSM attacks, consistently requiring the maximum allowed iterations ($30$) on both datasets. For PGD attacks (Fig.~\ref{PGD2} and \ref{PGD2-100}), our model shows slower attack speed at small $\epsilon$. These results indicate lower counterfactual generating speed and support our model's enhanced interpretability.

The ROAR results presented in Table~\ref{roar} provide additional validation of our model's interpretability. Higher ROAR scores indicate that removing important features significantly impacts model accuracy, suggesting better alignment between the model's decision-making process and attribution methods. Our model achieves the highest ROAR among all compared approaches, demonstrating that saliency maps more accurately reflect its interpretability. The consistently lower ROAR of discriminative models compared to generative approaches underscore the interpretability advantages of generative architectures. The comprehensive evaluation results reveal a clear correlation between adversarial robustness (discussed in Section~\ref{robustness}) and model interpretability. Models displaying greater vulnerability to adversarial attacks consistently demonstrate lower interpretability across our metrics. This relationship suggests that our model's superior adversarial robustness stems from its enhanced interpretability, highlighting the importance of interpretable architectures in developing robust DNN models.

\begin{table}[h!]
  \caption{ROAR. Higher ROAR shows more accuracy degradation when important features are missing, indicating better interpretability.}
  \centering
  \renewcommand{\arraystretch}{1}
  \begin{tabular}{|p{4cm}|l|}
    \hline
    \textbf{Network}     & \textbf{ROAR} \\
    \hline
    VGG16 \cite{simonyan2014very} &   0.4828      \\
    DBX-FC1 \cite{li2019generative} & 0.4962   \\
    \hline
    Deep CAMA \cite{zhang2020causal} &  0.2462             \\
    GBZ-CONV9 \cite{li2019generative}     & 0.4278  \\
    GBY-CONV9 \cite{li2019generative} &0.5571    \\
    \hline
    Ours (deep)    & 0.5393  \\
    Ours (shallow)    & \textbf{0.7405}  \\
    \hline
  \end{tabular}

  \label{roar}
\end{table}

\begin{table}[h!]
  \caption{Ablation study on various layers of discriminative features.}
  \centering
  \renewcommand{\arraystretch}{1}
  \begin{tabular}{|l|l|c|}
    \hline
    \textbf{Dataset} & \textbf{Model}    & \textbf{Acc} \\
    \hline
    \multirow{2}{*}{CIFAR-10}&Shallow (9 CONV)  & 0.9248   \\
    
    &Deep (13 CONV+1 Dense) &  0.9304             \\
    \hline
    \multirow{2}{*}{CIFAR-100}&Shallow (12 CONV) &0.7011    \\
    &Deep (13 CONV+1 Dense)    & 0.7089  \\
    \hline
     \multirow{2}{*}{\makecell[l]{Tiny-\\ImageNet}}    &Shallow (14 CONV)   & 0.4734  \\
    &Deep (16 CONV+2 Dense)    & 0.6935  \\
    \hline
  \end{tabular}
  \label{abl}
\end{table}

\subsection{Ablation Study}
\label{ablation}

We conduct an ablation study to investigate the impact of combining discriminative features and generative classifiers in our model architecture. Specifically, we examine how the depth of the feature extractor affects various aspects of model performance. By varying the number of layers in the feature extractor, we uncover important trade-offs between clean accuracy and adversarial robustness. The experimental results presented in Table~\ref{abl} demonstrate that deeper feature extractors lead to higher clean accuracy, attributable to their ability to learn more sophisticated discriminative features. However, as shown in Figure~\ref{5} and Table~\ref{roar}, models with shallower feature extractors show superior robustness and better interpretability. This observation points to a fundamental trade-off in our architecture: more sophisticated discriminative features, while beneficial for clean accuracy, tend to compromise adversarial robustness. The relationship suggests that achieving optimal performance requires careful balancing of the feature extractor's depth.

To validate the generalizability of our approach, we extend our experiments to the more challenging Tiny-ImageNet dataset using a VGG19 backbone. While shallow models struggle to achieve satisfactory performance with clean accuracy below $0.5$, deep models demonstrate SOTA performance with accuracy around $0.7$. Based on previous findings, we argue that with appropriate architectural choices, our approach can be extended to develop Tiny-ImageNet models that achieve both acceptable accuracy and adversarial robustness.

\section{Conclusion}
\label{5}

In this paper, we present a generalized image classifier architecture that combines discriminative features and generative classifiers with built-in causal graphs to achieve both high classification accuracy and adversarial robustness. The experimental results on CIFAR-10 and CIFAR-100 datasets validate our model's superior adversarial robustness compared to SOTA adversarially trained generative classifiers. Notably, our model achieves this robustness while requiring no adversarial example augmentation during training. Through interpretability analysis using multiple evaluation metrics, our results reveal strong correlations between model interpretability and adversarial robustness, suggesting that enhanced interpretability contributes to improved robustness. The flexibility of our architecture is demonstrated through its extension to the more complex Tiny-ImageNet dataset. The generative component functions as an auxiliary network that can be integrated with various pre-trained CNNs, adapting to different dataset requirements. While our current work focuses primarily on white-box robustness, future research can include evaluating our approach against black-box attacks , and extending our methodology to other computer vision tasks such as object detection.


\section*{Acknowledgments}
This work was partially supported by the Center for Connected Multimodal Mobility ($C^2M^2$) (U.S. DOT Tier 1 University Transportation Center) headquartered at Clemson University, Clemson, South Carolina.

\bibliographystyle{elsarticle-num}
\bibliography{main}

\end{document}